\documentclass[letterpaper, 10 pt, conference]{ieeeconf}  

\IEEEoverridecommandlockouts                              

\usepackage{cite}
\usepackage{amsmath,amssymb,amsfonts}
\usepackage{algorithmic}
\usepackage{graphicx}
\usepackage{textcomp}
\usepackage{xcolor}
\usepackage{hyperref}

\title{\LARGE \bf
Mining multi-modal communication patterns in interaction with explainable and non-explainable robots 
}

\author{Suna Bensch$^{1}$, Amanda Eriksson$^{1}$
\thanks{$^{1}$Department of Computing Science,
        Ume{\aa} University, 901 87 Ume{\aa}, Sweden.
        {\tt\small suna@cs.umu.se, amer0028@student.umu.se}}%
}

\begin{document}

\maketitle
\thispagestyle{empty}
\pagestyle{empty}

\begin{abstract}
We investigate interaction patterns for humans interacting with explainable and non-explainable robots. Non-explainable robots are here robots that do not explain their actions or non-actions, neither do they give any other feedback during interaction, in contrast to explainable robots. 
We video recorded and analyzed human behavior during a board game, where 20~humans 
 verbally instructed either an explainable or non-explainable Pepper robot to move objects on the board. The transcriptions and annotations of the videos were transformed into transactions for association rule mining. Association rules discovered communication patterns in the interaction between the robots and the humans, and the most interesting  rules were also tested with regular chi-square tests. Some statistically significant results are that there is a strong correlation between men and non-explainable robots and women and explainable robots, and that humans mirror some of the robot's modality. Our results also show that it is important to contextualize human interaction patterns, and that  this can be easily  done using association rules as an investigative tool. The presented results are important when designing robots that should adapt their behavior to become understandable for the interacting humans.
\end{abstract}

\section{INTRODUCTION}
There is an increasing interest in explainability for AI in general~\cite{calvaresi2022}. In robotics research, explainability goes also under several other names, for example understandability~\cite{hellstrom2018understandable}, readability~\cite{taka11}, anticipation~\cite{giel11}, legibility~\cite{licht14,licht12}, and predictability~\cite{Daut05}. The specific focus also varies a lot and includes how robots should be designed such that humans understand their actions~\cite{SinghBaranwalRichterHellstromBensch2021} or explanations~\cite{MUALLA2022103573}, motions~\cite{Zhou17, Dragan13, cakl15}, emotions~\cite{ky09, ksbwkb13}, goals~\cite{kaaa17}, failures~\cite{raman2013}, and  various types of social behavior~\cite{matarese2021, Breazeal02, Dautenhahn98}.
The benefits with explainable robots has also been investigated in several studies.
In~\cite{paleja2021}, explainability was shown to support situational awareness in human-machine teams. In~\cite{nk11}, robots acting without communicating their intentions to involved humans created anxiety. In cooperative tasks, the  efficiency has been shown to be negatively affected if the human cannot correctly predict the actions of the robot~\cite{giel11, bmnss14}. 
Despite this acknowledged importance of, and active research in, explainable robots, there is still little knowledge about human communication patterns in interaction with explainable and non-explainable robots.  Such knowledge about communication patterns or behavior is important when designing explainable robots that should adapt their behavior in order to increase explainability for the interacting human.

In this paper, we investigate human communication patterns when interacting with explainable and non-explainable robots using association rule mining. 

Our results show that humans mirrored the robot's choice of modality, and that there was a strong correlation between male users and non-explainable robots and female users and explainable robots. Male user tended to engage verbally and with gestures when interacting with non-explainable robots, whereas female users engaged verbally and with gestures when interacting with explainable robots. Typical examples for such interactions are shown in Table~\ref{dialogues}.

\begin{table}[htbp]
\caption{Example dialogues for men reacting to non-understandable robots and women reaction to understandable robots.}
\begin{center}
\begin{tabular}{|l|l|}
\hline
\textbf{Agent} & \textbf{Utterance} \\ \hline
Man: &  Can you move the yellow triangle to square 11?\\
Robot: &  No. \\
Man: & No? OK. \\ \hline
Woman &  What about moving it to Cell 9? \\
Robot & I cannot reach Cell 9. You have to find another way. \\
Woman &  Oho, right. \\  \hline
\end{tabular}
\label{dialogues}
\end{center}
\end{table}

Note here that the variable \emph{gender} emerged as a significant variable through the generated association rules (it is not a controlled independent variable).

We annotated and analyzed 20 videos (amounting to around 200 minutes) of human-robot interactions, where the participants played a game with either an explainable or non-explainable robot, instructing the robot to move certain pieces on a virtual game board displayed on the chest screen of a Pepper robot. The annotations were translated into 295~\emph{transactions} that were used to generate association rules describing relations between several observed variables (e.g. both human and robot verbal utterances, gestures, no feedback). 
 This approach allowed us to investigate multi-modal communication patterns in interaction with explainable and non-explainable robots, and how the observed variables co-varied, and possibly affected each other. 
The usage of association rules as an investigative tool demonstrated the importance of contextualizing human behavior and feedback. For example, we found that regular statistical tests of general questions, such as \emph{Is there a difference in how humans behave when interacting with understandable versus non-understandable robots?}, lead to the conclusion that no difference can be shown. In terms of association rules, the Lift for these rules was approx. 1.0, meaning that \emph{explainability} and \emph{human behavior} were independent events. However, when more context was taken into account, which corresponds to more items in the antecedent (explained in detail below), differences in robot explainability and gender emerged. 
Such contextualization is important for HRI research, in particular for design of explainable robots, that should interpret human behavior as accurately as possible. For example, our results indicate also that absence of human feedback not necessarily is a signal of non-understanding. 

\section{METHODS}
\subsection{Experimental and robot setup}
\begin{figure}[htbp]
\centerline{\includegraphics[scale=0.35]{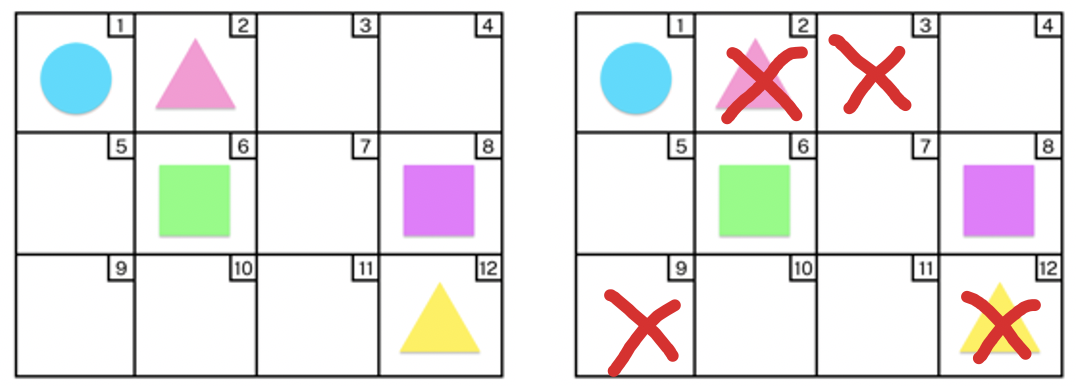}}
\caption{To the left is the initial configuration of the game board, which was displayed on the robot's chest-mounted tablet. The game was won when the blue circle was moved to Cell~4. Cells 9 and 3 were not reachable and the triangles were not movable as illustrated to the right.}
\label{Image1}
\end{figure}

We target a scenario where a human plays a board game with a Pepper robot, verbally instructing the robot to move pieces on the board with the goal to move a certain piece to a specified cell (see Fig.~\ref{Image1}). 
Some pieces on the board were not movable and some cells were not reachable, however the human had no knowledge of which pieces or cells these were but had to find out in interaction with the robot. 
The human was either interacting with an explainable or non-explainable robot. An explainable robot, for example, explained why it could not execute an invalid instruction given by the human (e.g \emph{Human: Move the green square to Cell 3. Robot: I cannot reach Cell 3}), whereas a non-explainable robot did not execute an invalid instruction without giving the human an explanation or gesture feedback. 

Twenty users, thereof seven identifying as female and thirteen identifying as male, interacted with either an explainable or non-explainable robot. The age range was between 20 and 45, sixteen of the users were students studying computer science, interaction design, physics, or engineering. 
The goal for each user was to verbally guide the robot to move the blue circle from Cell~1 to Cell~4 (see Fig.~\ref{Image1}) by giving the robot a sequence of instructions on objects to move. 
The users were informed that some of the objects were not movable and some cells were not reachable for the robot and they had to find out which objects and cells these were by interacting with the robot and exploring possible paths to reach the goal.  

The robot was remote-controlled by a Wizard who sat in an adjacent but separate room using the WoZ4U interface~\cite{Rietz2021}. 
The behavior of the explainable robot included explaining when and why it could not execute the human's instruction, for example, \emph{I cannot reach Cell~3, you have to find another way} or \emph{I cannot move the triangle}. 
The behavior of the non-explainable robot was less verbal or expressive in gestures, even though the robot sometimes uttered \emph{yes} or \emph{no}, and nodded or shook its head to signal, for example, invalid instructions or auditory problems. 
In most cases, the role of the non-explainable robot was to not react, to give no feedback at all. The robot would, for example, often do nothing at all when invalid instructions were given by the human, or when the robot (the Wizard) had auditory problems and could not hear what the participant said. 

\subsection{Video annotation and analysis}\label{subsec:D}
\begin{figure}[htbp]
\centerline{\includegraphics[scale=0.3]{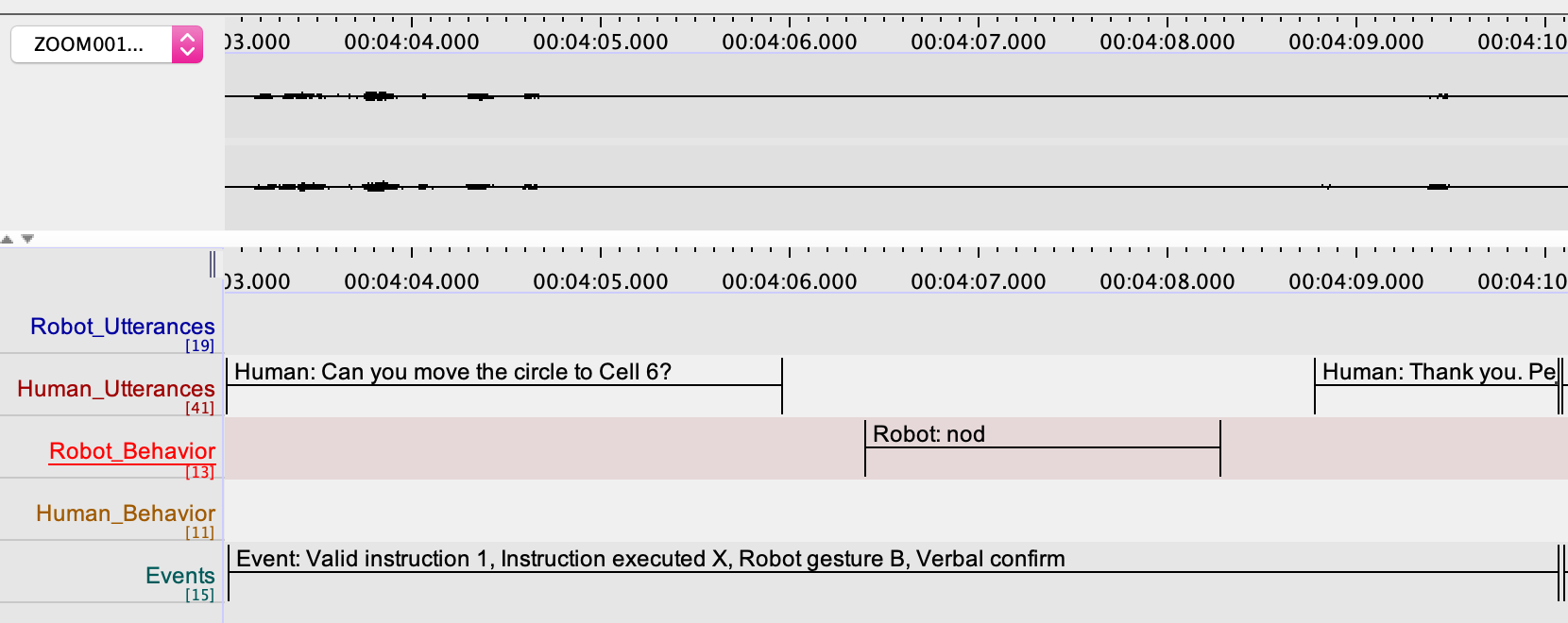}}
\caption{Sample excerpt of a transcription and initial annotation in ELAN of an interaction between a non-explainable robot and a user.}
\label{ELAN_picture}
\end{figure}

Each participant interacted with the robot for, in average, around 10~minutes. The interactions with the understandable robot were in average 9.7~minutes long, whereas the interactions with the non-understandable robot were in average 11.6~minutes. 
The verbal and non-verbal interactions between the humans and the robot were manually transcribed and annotated with the video annotation software ELAN~\cite{ELAN} by one annotator (see Figure~\ref{ELAN_picture}). 
In addition, we defined the notion of \emph{events}. An event starts with a participant's instruction and ends just before the participant's next instruction begins. 
An event consists of all human and robot verbal utterances and behaviors (i.e. gestures or no reaction), information on whether the human instruction was valid or not, and whether the instruction was executed or not).

\subsection{Association rules and dataset}
Association rule mining introduced in~\cite{10.1145/170036.170072} is today an umbrella term for algorithms that identify associations for co-occurring instances in a dataset. The terminology necessary for this paper is described in the following~\cite{10.1145/170036.170072}.

Let ${\cal{I}} =\{I_1, I_2, \ldots, I_n \}$ be a set of binary attributes called \emph{items}.
Let $T = \{t_1, t_2, \ldots, t_m\}$ be a database of transactions. Each transaction $t$ is represented as a binary vector where each element is true or false depending on whether the corresponding item belongs to the transaction or not. 
An association rule is written as: 
$\alpha \to \beta$ where $\alpha, \beta \subseteq {\cal{I}}$ and $\alpha \neq \beta$. Both $\alpha$ and $\beta$ are referred  to as \textit{item sets}.
We refer to the left hand side $\alpha$ as \emph{antecedent} and to the right hand side $\beta$ as \emph{consequent}. The interpretation of an association rule is ``the observation of the antecedent $\alpha$ is correlated with the observation of the consequent $\beta$".

Given a dataset of transactions $T$, association rules are generated 
satisfying certain conditions: 
\subsubsection{Syntactic constraints} are specifications of the exact items that should appear in the antecedent and consequent.  
\subsubsection{Support} is the number of transactions containing $\alpha$ and $\beta$, divided by the total number of transactions in $T$. Support is a measure of how frequently a rule's item sets occur in a data set. 
\subsubsection{Confidence}  is defined as the support of $\alpha$ and $\beta$ divided by the support of $\alpha$. Confidence is a measure of how \emph{strong} a rule is, in the sense how strongly the occurrence of the antecedent is correlated with the the occurrence of the consequent. 
\subsubsection{Lift} is defined as the support of union of item sets in the antecedent and consequent divided by support of the antecedent times the support of the consequent~\cite{tan2005}. If Lift equals 1, the antecedent and consequent are independent. If Lift is $>1$, there is a positive correlation between the antecedent and consequent and they appear more often together than they would if they were independent. If Lift is $<1$, there is a negative correlation between the antecedent and consequent and they appear less often together than they would if they were independent. 
Hence, while Confidence for a rule may be high simply because the consequent in general is very common in data,  Lift compensates for this.

\begin{figure}[htbp]
\centerline{\includegraphics[scale=0.23]{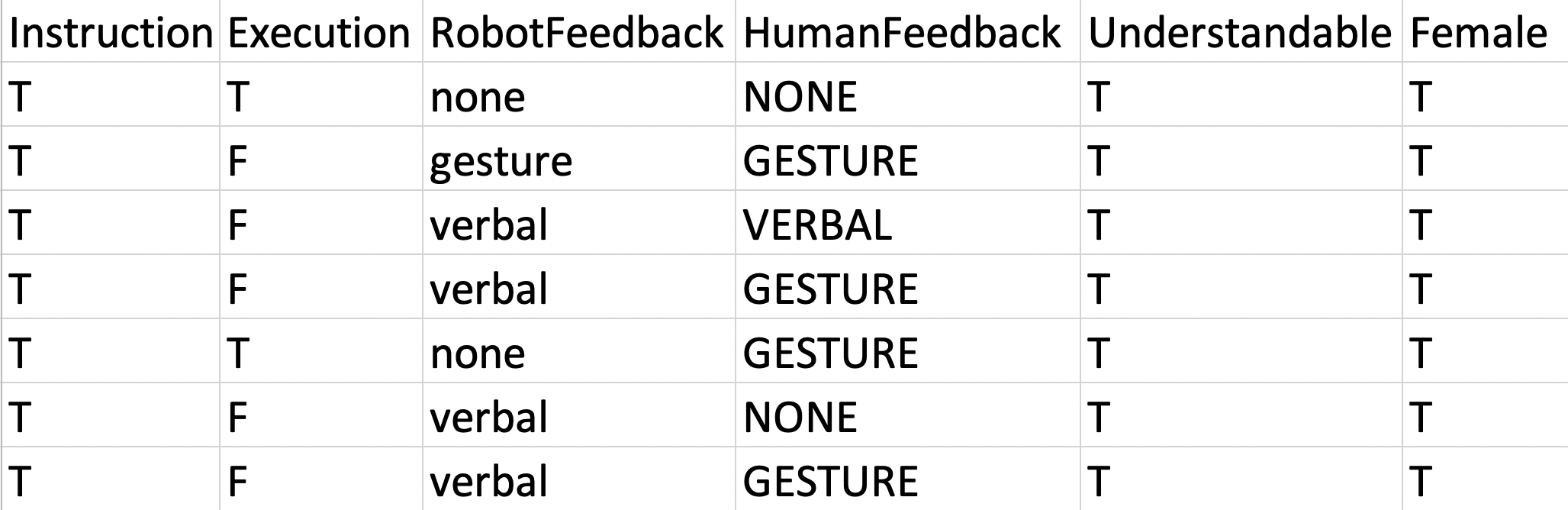}}
\caption{Excerpt from the in total 295 transactions. Each row is a compressed representation of a transaction. For compact representation, the items are not only binary but take discrete values.}
\label{Transactions}
\end{figure}

All events were transformed into 295 \emph{transactions} for association rule mining as shown in Figure~\ref{Transactions}
and Table~\ref{data} shows item names and the possible values, where {\tt T} stands for {\tt TRUE} and {\tt F} for {\tt FALSE}. 

\begin{table}[htbp]
\caption{Items used for generation of association rules}
\begin{center}
\begin{tabular}{|l|l|l|}
\hline
\textbf{Item name} & \textbf{Values} \\ \hline
Execution &  \{T, F\} \\
Instruction & \{T, F\} \\
RobotFeedback & \{verbal, gesture, none\} \\
HumanFeedback & \{VERBAL, GESTURE, NONE\} \\
Understandable & \{T, F\} \\
Female & \{T, F\}\\
\hline
\end{tabular}
\label{data}
\end{center}
\end{table}

To each transaction we added the information whether the robot was explainable or not (i.e. {\tt Understandable=T} or {\tt F}) and whether the user was female or male (i.e. {\tt Female=T} or {\tt F}) (see Fig.~\ref{Transactions}). 
Thus, a transaction contains information on whether or not an instruction was valid, whether or not the instruction was executed, the type of robot feedback, the type of human feedback, whether or not the robot was understandable, and whether or not the participant identified as female. 
 The data from one male participant, interacting with a non-understandable robot, was excluded since they decided to leave the experiment. The 295 transactions were  input to the Orange association rules mining tool~\cite{Orange}, which implements FP-growth frequent pattern mining algorithm~\cite{Han2004} with bucket optimization~\cite{Agrawal00}.
The minimum Support was set to 3\%, the minimum Minimum confidence level was set to 20\%, and the minimum Lift was set to $> 1.0$. These settings were a trade-off between  the number of transactions (monitored through the Support value) for each generated rule, and  the total number of interesting generated rules (monitored through the Lift parameter). 

Furthermore, we controlled the generation of association rules by specifying syntactic constraints for the antecedent and consequent, i.e. which items to include in the rules. The consequent was constrained to be {\tt HumanFeedback=VERBAL}, {\tt HumanFeedback=GESTURE}, or {\tt HumanFeedback=NONE}.
By not specifying any items in the antecedent, all items were allowed. 


\section{RESULTS}\label{sec:results}
\begin{figure}[htbp]
\centerline{\includegraphics[scale=0.3]{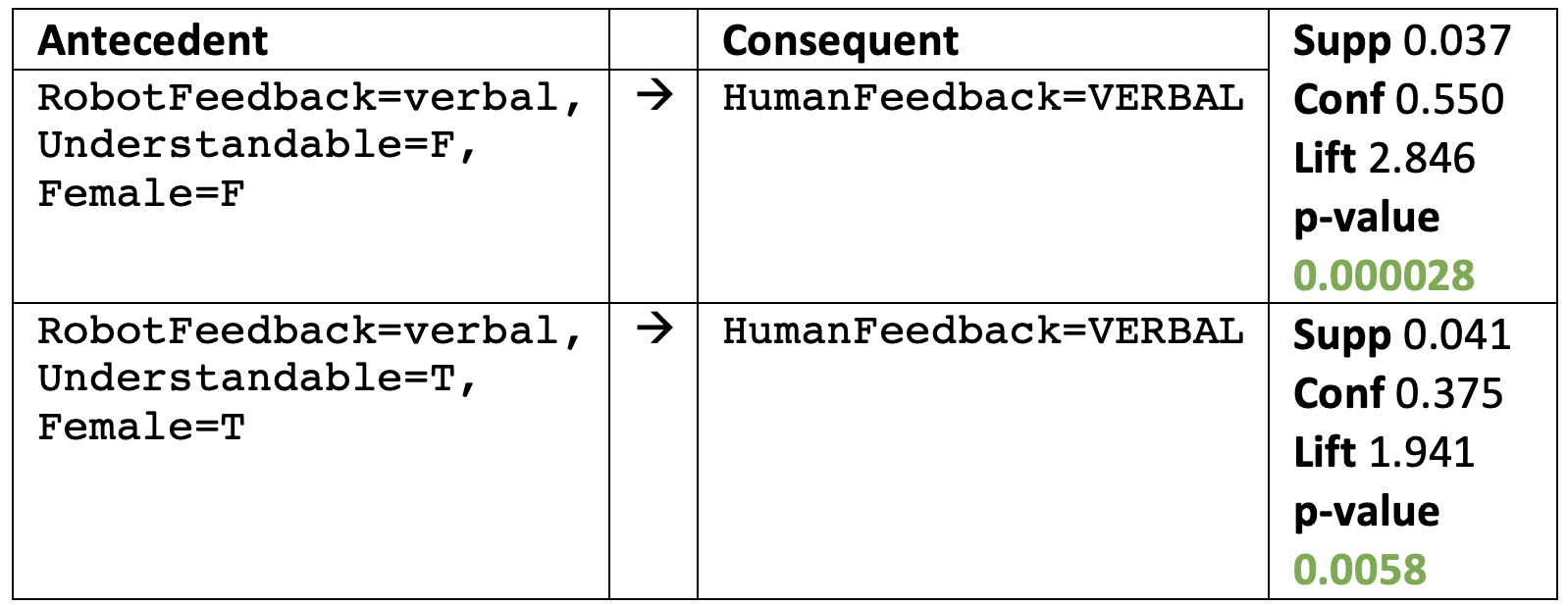}}
\caption{Generated association rules with  Lift $\geq 1.9$, indicating a gender effect with strong correlation between male users and non-explainable robots and female users and explainable robots, both using verbal utterances.}
\label{ResultsQ1}
\end{figure}

\begin{figure}[htbp]
\centerline{\includegraphics[scale=0.3]{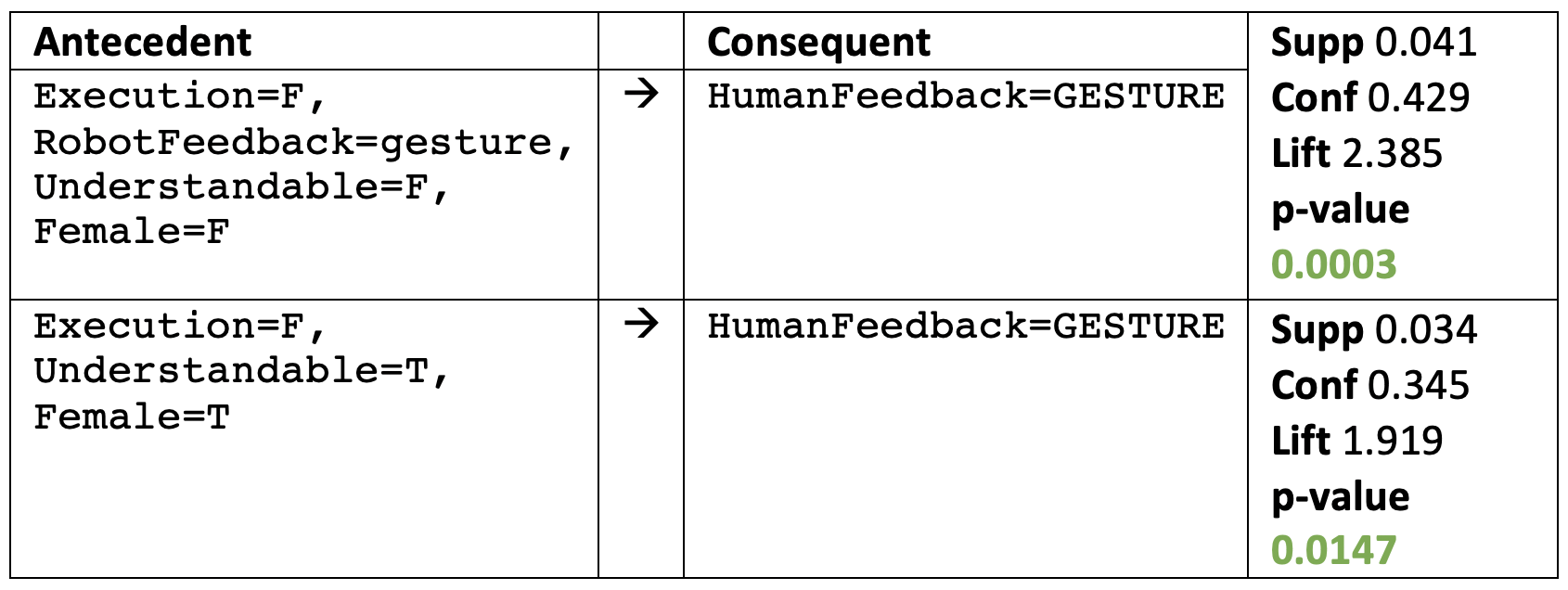}}
\caption{Generated association rules with Lift $\geq 1.9$, indicating a gender effect when explainable and non-explainable robots do not execute a human instruction.}
\label{ResultsQ2}
\end{figure}

\begin{figure}[htbp]
\centerline{\includegraphics[scale=0.29]{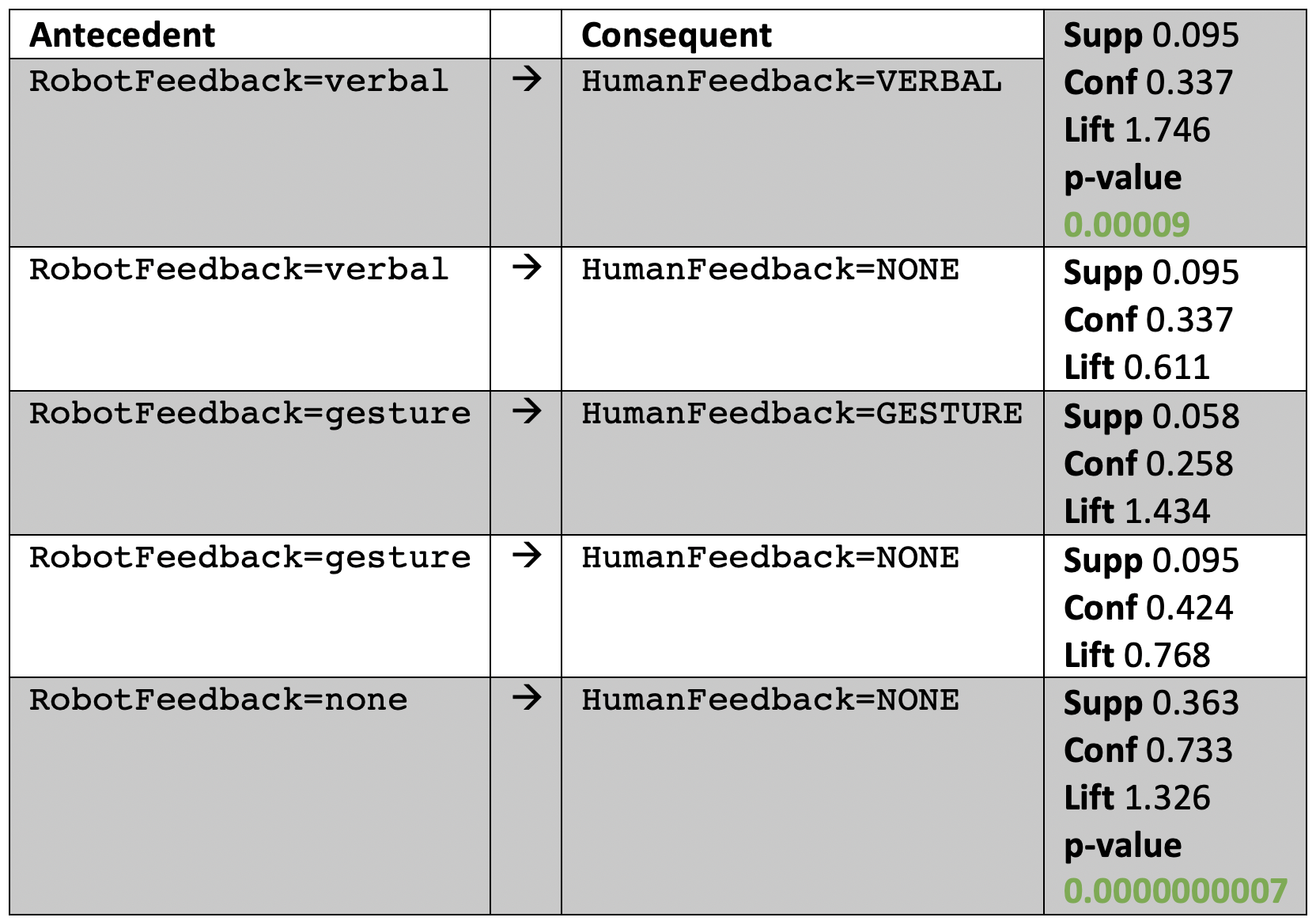}}
\caption{Generated association rules indicating that humans mimic the robot's modality. The third association rule with Lift~1.434 and Support~0.058 had no statistical significance.}
\label{ResultsQ5}
\end{figure}

\begin{figure}[htbp]
\centerline{\includegraphics[scale=0.3]{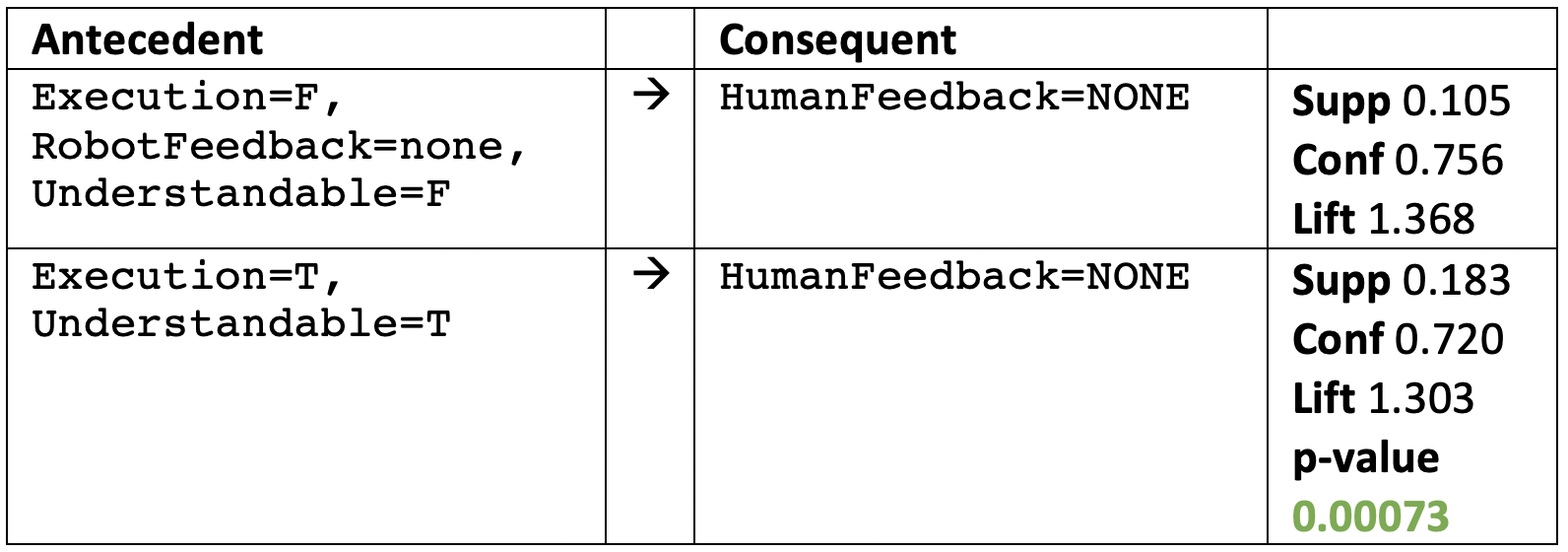}}
\caption{The first rule, even though not statistically significant, suggests that humans do not provide feedback when interacting with a non-explainable robot that does not execute a human instruction and does not give any robot feedback. The second rule shows that humans do not provide any feedback in interaction with an explainable robot executing the human instruction.}
\label{ResultsQ4}
\end{figure}

In this section we present some of the most interesting rules, that are both statistically significant and interpretable. The rules are described in Figures~\ref{ResultsQ1}-\ref{ResultsQ4}, showing the attributes in the antecedent as well as the consequent side. The Support, Confidence, and Lift values are also listed. To assess the statistical significance level of the dependence between antecedent and consequent~\cite{baxter2016}, p-values for conducted chi-square tests are also provided (for rules with $p<0.015$). 
Figure~\ref{ResultsQ1} shows two rules indicating gender effects. 
The first rule, with Lift~2.846,  indicates that male users who interacted with a non-explainable robot that gave verbal feedback, provided verbal feedback (rather than other types of feedback).
The second rule, with Lift~1.941,  indicates that female users who interacted with an explainable robot that gave verbal feedback, provided verbal feedback. 

The rules in Figure~\ref{ResultsQ2} indicate other gender effects, for cases where the robot did not execute an instruction given by the user. Here, female users who interacted with explainable robots, as well as male users interacting with non-understandable robots, tended to give gesture feedback. 

Figure~\ref{ResultsQ5} shows association rules suggesting that humans mimic the robot's choice of modality. The non-shaded rules have   Lift~$<< 1$. This means that there is a negative correlation between the antecedent and consequent in the transactions.

The  association rules in Figure~\ref{ResultsQ4} have the consequent {\tt HumanFeedback=NONE}, which stands for no  human feedback. 
The statistically significant second rule shows that humans provide no feedback to an explainable robot that executes an instruction. The first rule suggests that humans provide no feedback to a non-explainable robot that does not execute an instruction and provides no feedback itself. 


Our results show several statistically significant multi-modal communication patterns in humans' interaction with explainable and non-explainable robots:
\begin{enumerate}
    \item Men react verbally to non-explainable verbal robots, whereas women react verbally to explainable verbal robots.
    \item In cases where robots do not execute the human's instruction, men provide gesture feedback to gesturing non-explainable robots, whereas women provide gesture feedback to explainable robots.
    \item Humans provide no feedback when interacting with an explainable robot that executes an instruction. 
    \item Humans give verbal feedback if the robot does the same, and give no feedback if the robot does not give any feedback. 
\end{enumerate}

\section{DISCUSSION}
\begin{figure}[htbp]
\centerline{\includegraphics[scale=0.46]{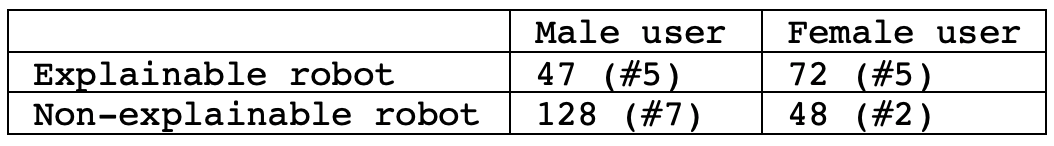}}
\caption{Number of transactions for male and female users interacting with explainable and non-explainable robots. The number of participants $\#$ is given in parenthesis.}
\label{NumberOf}
\end{figure}

Figure~\ref{NumberOf} lists the number of transactions for male and female users interacting with explainable and non-explainable robots. The number of users is given in parenthesis. The anonymized data can be provided by the first author upon request.  

Association rule mining provides several advantages for data analysis in HRI. One advantage is that interaction patterns can be investigated without specific hypotheses being formulated beforehand. Another advantage is that association rules allow for analysis of contextualized human behaviors by means of implication rules {\tt if} $\alpha$ {\tt then} $\beta$ for which further syntactic constraints can be given. 
For example, even if there were no indications of general differences between explainable and non-explainable robots and human behavior, statistically significant  association rules could be found when more contextual factors were taken into account. 

\section*{ACKNOWLEDGMENT}
The authors thank Thomas Hellstr\"om for suggesting the chi-square tests to assess the statistical significance of the rules. 

\bibliographystyle{IEEEtran}
\bibliography{biball}

\end{document}